\documentclass[conference]{IEEEtran}
\IEEEoverridecommandlockouts
\usepackage{cite}
\usepackage{amsmath,amssymb,amsfonts}
\usepackage{algorithmic}
\usepackage{graphicx}
\usepackage{textcomp}
\usepackage{xcolor}
\def\BibTeX{{\rm B\kern-.05em{\sc i\kern-.025em b}\kern-.08em
    T\kern-.1667em\lower.7ex\hbox{E}\kern-.125emX}}
\usepackage{subcaption}
\usepackage{amssymb}
\usepackage{pifont}
\newcommand{\cmark}{\ding{51}}%
\newcommand{\xmark}{\ding{55}}%

\begin{document}

\newcommand{\name}{PIETS}

\title{PIETS: Parallelised Irregularity Encoders for Forecasting with Heterogeneous Time-Series}

 \author{\IEEEauthorblockN{Futoon M. Abushaqra\IEEEauthorrefmark{1},
 Hao Xue\IEEEauthorrefmark{2},  Yongli Ren\IEEEauthorrefmark{3} and
Flora D. Salim\IEEEauthorrefmark{4}}
 \IEEEauthorblockA{School of Computing Technologies,  RMIT University\\
Melbourne, VIC, Australia \\
Email: \IEEEauthorrefmark{1}futoon.abu.shaqra@student.rmit.edu.au,
 \IEEEauthorrefmark{2}hao.xue@rmit.edu.au,
 \IEEEauthorrefmark{3}yongli.ren@rmit.edu.au,
\IEEEauthorrefmark{4}flora.salim@rmit.edu.au}}
\maketitle

\begin{abstract}
Heterogeneity and irregularity of multi-source data sets present a significant challenge to time-series analysis. In the literature, the fusion of multi-source time-series has been achieved either by using ensemble learning models which ignore temporal patterns and correlation within features or by defining a fixed-size window to select specific parts of the data sets. On the other hand, many studies have shown major improvement to handle the irregularity of time-series, yet none of these studies has been applied to multi-source data. In this work, we design a novel architecture, PIETS, to model heterogeneous time-series. PIETS has the following characteristics: (1) irregularity encoders for multi-source samples that can leverage all available information and accelerate the convergence of the model; (2) parallelised neural networks to enable flexibility and avoid information overwhelming; and (3) attention mechanism that highlights different information and gives high importance to the most related data. Through extensive experiments on real-world data sets related to COVID-19, we show that the proposed architecture is able to effectively model heterogeneous temporal data and outperforms other state-of-the-art approaches in the prediction task.
\end{abstract}

\begin{IEEEkeywords}
heterogeneous data sources, irregular time-series, sequences modeling, temporal data analysis.
\end{IEEEkeywords}

\section{Introduction}
Modeling multi-source time-series is a critical process that requires handling several issues related to the irregularity and inconsistency of heterogeneous data sets. Using multi-source data is essential for many real-world  applications including urban systems, internet-of-things, healthcare, and smart cities applications ~\cite{horn2020set,martinez2015survey,lim2020time}. It provides a significant improvement over uni-source systems, especially for tasks that are usually caused and affected by many complex factors. Suicide and crime event prediction,  early diagnosis of rare diseases, and disaster management are examples of such tasks, where the required information originates from multiple applications, tools, or organizations.  For instance, COVID-19 forecasting has been recently addressed as a hard prediction task that requires a huge amount of information from many sectors. Several studies have addressed multi-source data for COVID-19 prediction and data sets related to  mobility, health, weather and geographical information was used \cite{mahalle2020data, DBLP:journals/sncs/ShindeKMDCH20, DBLP:journals/jms/Santosh20}.

However, multi-source time-series analysis is highly challenging due to the irregularity of the data in terms of observation time and sequence lengths \cite{shukla2021multi}. Therefore, it is unable to be processed using traditional machine learning that requires samples captured at the same time with fully-observed fixed-size features and is incapable to capture temporal features. Contrastingly, Deep Learning (DL) sequence models such as RNN, LSTM, and GRU showed great performance for modeling sequence data and handling samples with different lengths. However, RNN models assume equally-sized intervals between observations \cite{sherstinsky2020fundamentals}.

Heterogeneous time-series modeling is a new issue and has not been explicitly considered previously. Most of the earliest studies on irregular temporal data either ignored the temporal information and time aspect or used selected sub-sequences from the data based on a predefined window, which leads to losing a lot of information. Recently, several techniques have been presented to handle this issue and model irregularity time-series. Some of these techniques focus on pre-processing the data to be suitable for existing models. Such studies have either used imputation or feature extraction and representation methods \cite{pratap2019multi ,che2018recurrent,kim2018temporal,zhang2021feature}. On the other hand, novel studies presented models that were designed to handle irregularity e.g. RNN-decay, ODE-based models, and CDE-based model \cite{de2019gru,rubanova2019latent}. These models were designed to have a continuous hidden state which provides flexible models to represent sequences with non-uniform intervals. However, pre-processing the data and applying imputation are not effective methods for multi-source data as they produce unreliable observations and are limited to a small number of missing values.  In addition,  recent irregularity models are not efficient for highly sporadic time-series and have not been designed for multi-source data. One study addressed the multi-modality time-series \cite
{horn2020set}, the authors represented the each observation as a tuple of (time, value and modality) and aggregated all observations from several modalities using set function. To bridge the gap of handling multi-source inconsistent time-series data set, several difficulties should be discussed and addressed including:

\textbf{Heterogeneity and irregularity:} Irregular sampling occurs in several time-series modeling and presents a substantial challenge.  Irregular data defined as time sequences with non-uniform intervals i.e. multivariant data where variables differ in terms of observation time and frequency. However, having heterogeneous data sources emphatically leads to irregular samples, as each data set covers different periods with different starting (initial) time points. The sequences extracted from heterogeneous sources are highly sporadic and have non-unified lengths with a large amount of missing observation. Accordingly, modeling multi-source sequences require an architecture that leverages all the captured data from non-uniform time points.

\textbf{Information inconsistency:} Combining data from many sources produces a huge amount of information whereas the knowledge is derived from different tools and applications. Therefore, the types of attributes will be inconsistent as each data set has a different representation, distribution, scale, and density. For example, features extracted from health applications are different from features extracted from weather data sets. This considers a major issue since the inconsistent data is hard to be modeled and fused using traditional methods.

\textbf{Extremely large and highly-variable dimensions:} 
Another key challenge of using several data set is the highly-variable dimensions of information. Including several data sources produce a massive amount of correlations and temporal patterns that might be hard to be handled by a single model. Besides, each set of features has a different impact and effect on the target value.  For example, if a specific data source has a strong correlation and a high impact on the task; other data sources might have less effect and are only used to emphasize the result for a specific case. Defining the importance of each set of features is an ambiguous task as it  differs from one sample to another and required extensive experience.

In this work, we investigate how to model multi-source time-series and focus on handling the issues of irregularity,  unequal-length sequences, inconsistency and high-dimensionality data. We propose a new neural network architecture (PIETS) for modeling time-series data from heterogeneous sources with three main contributions:
\begin{itemize}
\item We design irregularity encoders to handle both the irregular series captured from heterogeneous data sets and the successive missing values that occur in different feature sets. This encoder leverages information of all observations rather than eliminating any sub-sequence or add unreliable imputed values.
\item We propose parallelised sources modeling to preserve the ability to represent temporal patterns between observations from each source and give the architecture the flexibility to independently learn the correlation between inconsistent data sets and target values. The parallelised modeling helps to process the inconsistent features that have different attributes.
\item We introduce an attention-based feature fusion mechanism, which gives the ability to discover any correlations between data from all sources as well as highlights and focuses on different parts of information particularly the most valuable parts, attention-based module can manage the massive amount of data and prevent overwhelming.
\end{itemize}

 

\section{Related Work}
Many recent studies have been presented to solve the challenges associated with time-series data, including missing values \cite{cao2018brits,liu2019naomi,ma2020midia,khayati2020mind}, features extraction \cite{zhang2021feature}, irregularity and multi-resolution \cite{pratap2019multi,li2016scalable,shukla2021multi,kidger2020neural}. In this article, we are interested in a new issue related to the multi-source time-series data which has not been explicitly considered previously. 
Therefore, in this section, we describe the major previous works that are related to irregularity and fusion of time-series data.

\subsection{Irregularity Time-series}
Several articles have discussed the irregularity of time-series data, especially for Internet of Health (IOH) applications, where despite that all information coming from one source (e.g. ICU), the captured features are recorded sparsely and irregularly \cite{lipton2016directly,zhang2021feature}. This characteristic is contrary to the machine learning models that assume a fixed-size of features \cite{yadav2018mining}. Consequently, most of the earliest studies on irregular temporal data either ignored the temporal information and time aspect or used selected sub-sequences from the data based on a predefined window. Using these solutions a lot of information and samples will be lost. 

Another way for dealing with the irregularity in time-series data is to impute and interpolate the missing samples \cite{pratap2019multi,che2018recurrent}. Many types of imputation methods could be applied such as the traditional simple statistical methods (minimum, maximum, mean, zeros), or more complex methods like Gaussian processes, K-NN and linear regression \cite{moor2019early, aittokallio2010dealing,zhang2008parimputation,raghunathan2001multivariate}.

However, these imputation methods cause the loss of temporal dependency. Particularly when applying to a huge amount of data where patterns and time intervals are important information. Hence, more effective methods have been presented. These recent methods consider the correlations of the variables by using observations without missing values to predict the missing values in other observations. Recently \cite{zhang2021feature} presented a novel method for the unequal-length Electronic Health Record (EHR) data and irregular samples. Their method used dynamic time warping to measure the similarity of temporal sequences thus find the alignment between sequence pairs (patients). The similarity matrices were stacked into a tensor. And tensor decomposition was used to extract the latent features and reduce the dimensionality of the data. In another work \cite{ma2020midia}, the authors improved an imputation method using the non-linear correlations between missing and available values, information about the missing patterns, and denoising auto-encoder (DAE). Moreover, rather than imputed the exact value of the missing observation, a recent work by Singh et al. \cite{pratap2019multi} built a model to impute the representation of the missing values. This model considered the correlation between available observation and other information related to missing value (such as the average value of the missing observation, the last observed value and the time lapse since the last observation). The framework was built using Bidirectional LSTM with Attention. The input for the model is a vector that includes the values of the feature at each time step along with the vector for a supporting signal. Each vector includes the mean, last observation, missing flag, and time-lapse. However, these imputation methods show good performance for data with few missing values but it is less effective when having sequences of missing data or variables with low correlation. 

Other sophisticated and graceful approaches have been used to handle time-series irregularity by providing more generalized models. Whereas recurrent neural networks that are widely used for modeling temporal data assume fixed laps between observations and fully observed samples \cite{sherstinsky2020fundamentals}, it shows less performance and unwieldy fitting for irregular samples. Many works used models based on ODE \cite{rubanova2019latent,chen2018neural} to deal with irregular time intervals and gaps in time-series data. ODE is used to define derivatives of the state with respect to time. Therefore it describes the evaluation in time. Neural ODE \cite{chen2018neural} has been presented as a  new family of deep neural network models. Recently, in \cite{rubanova2019latent} the authors provided an ODE-RNNs and Latent ODEs models that have a continuous-time latent state, where the formation of the dynamics between observations are not predefined, alternatively it is learned. Their methods work better with highly sparse data. Since that the neural ODE is not able to adjust the trajectory based on the new observations and its solution is determined by its initial condition, Kidger et al. \cite{kidger2020neural} tried to figure out how to incorporate incoming information. They proposed a neural controlled differential equation (Neural CDE) model that is able to process newly arrived data. Brouwer et al. provided GRU-ODE-Bayes \cite{de2019gru} which is a continuous-time Gated Recurrent Unit used to model the sporadic observations. Moreover, recent approaches used attention mechanism to model irregularly sampled time series \cite{shukla2021multi}.  Nevertheless, all the mentioned approaches have been tested and evaluated on uni-source data sets where the source of information was unified. 

Using multi-modality time-series has been addressed recently by  Horn et al. \cite{horn2020set} which proposed a set network approach for time-series prediction from non-synchronized sequences. The authors defined the non-synchronized time series as a time series with at least one time point where one modality did not make an observation. The proposed model receives inputs of time-series data from several modalities and all observation is represented as a tuple of time, value and modality, therefore that the time is encoded into the feature vector of points. All elements of each set are summarized using a set function $f$ with an attention mechanism to learn the importance of the individual observations.

\subsection{Data Fusion of Multivariate Time Series}
Using data from different modalities and sources requires a suitable and effective fusion technique. In general, fusion methods could be applied on several levels including data-level, feature-level, model-level, and decision-level \cite{wu2018sensor}. Data and feature levels are considered to be early fusion. With data-level methods, the raw data from all sources and modalities are combined. While using feature-level methods, the extracted features are combined in one vector. On the other hand, model and decision levels fusion are late fusion methods. Model-level fusion is used by concatenating different neural network layers before the final output. Finally, decision-level fusion is applied using ensemble techniques where different models produce the final outputs and several algorithms are used to estimate the last result.

Each of the fusion techniques works for different tasks. However, the fusion of multivariate time series data is more complicated due to the time effect. Most of the studies provided a multi-sensor data fusion \cite{mitchell2007multi,brena2020choosing} while recently the efforts expand to enhance these methods to handle the time-series features \cite{diao2019data,zhu2019improvement,diao2017data}. A common way is to use hierarchical architecture for fusion \cite{harris1998multi}. Also, other studies adopted ensemble learning without combining the data \cite{wu2018sensor}. The fusion of time-series data from different heterogeneous sources has not been well studied. While having ensemble models enables the system to handle the heterogeneity of the data it misses information and patterns related to the correlations between different features.

\section{Preliminaries}
\begin{itemize}
\item Observation ($O$): a record of data that has been captured at a specific time step ($i$), each observation includes a set of features ($D$) that describe a specific object.
\item Time-series data ($TS$):  a set of sequenced $N$ observations represented by $N \times D$ matrix  $TS = \left [t_{1},t_{2}, \cdots,t_{n} \right]$.  Where $t_{i} \in TS$ is a list of features $D$ that represent one observation at time $i$.  $i$ is the time from $1$ to $N$ with specific time intervals. 
\item Sub-sequence of a time-series ($TS_{i}^m$): a specific continuous subset of time series $TS$, starting from time instant $i$ until time instant $m-1$ with size $m$ . Such as $T_{_{i}m} = \left [t_{i},t_{i+1},t_{i+2},\cdots,t_{m-1} \right]$.
\item Overlapping sub-sequences:  a set of sub-sequences of a time series $TS$ extracted after applying a sliding window of length $m$. Where a sliding window with length $m$ create $n-m+1$ sub-sequences.
\item Data Source ($DS$):  the source from which a specific observation is coming. A data source $DS$ starts recording observation at initial time step $Ini$ with a time interval between each observation equal to ($p$). Multi-source time-series are data captured from several sources related to a specific task and used jointly to make predictions.  
\end{itemize}
Table \ref{table_notation} summarizes details about the symbols and notations used in the paper.
\begin{table}[!t]
\renewcommand{\arraystretch}{1.3}

\caption{Symbols and notations.}
\label{table_notation}
\centering

\begin{tabular}{|l||p{2.5in}|}
\hline
One & Two\\
\hline
$T$ & The timeline of all time frame.\\
$t$ & Specific time point on the time line $T$. \\
$D$ & Number of features describes specific observation for specific data source.\\
$S$ & Total number of data source used. \\
$Initial_{ts}$ & The initial time point $t$ where the source $s$ make the first observation. \\ 
$Present_{t}$ & the present time point on the timeline $T$. \\
$O_{ts}$ & Observation made at time point $t$ by source $s$.  \\
$EarlySub_{s}$ & Sub-sequences of data source $s$ starts from $Initial_{ts}$ and covers the Non-overlapping time-frame.\\
$LateSub_{s}$ & Sub-sequences of data source $s$ covers the Non-overlapping time-frame. \\\hline
\end{tabular}
\end{table}

\section{Methodology}
  \begin{figure*}[!t]
  \centering
  \includegraphics[width=\linewidth]{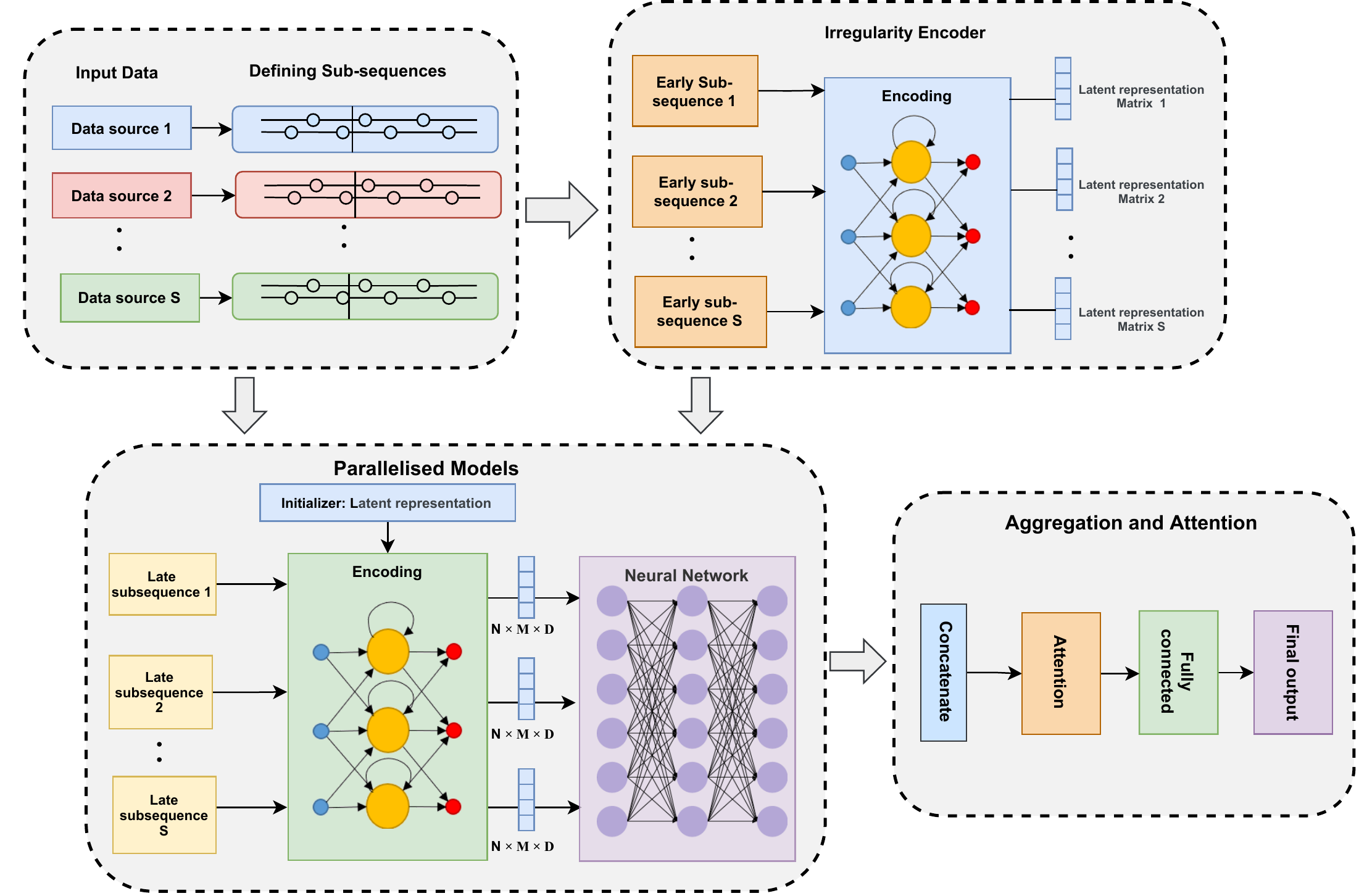}
  \caption{The framework of~\name\  consists of (1) irregularity encoder that producer pre-trained weights using the early sub-sequences data, (2) time-series parallelised models to encode late sub-sequences using corresponding weights matrix, (3) aggregating and attention module.  The encoder  produces $N \times M \times D$ matrix, where  $N$ is the number of samples, $M$ is the time point and $D$ is the dimensionality of the features. }
  \label{fig:two}
  \end{figure*}
  
\subsection{Problem Formulation}
We examine the problem of forecasting on $S$ time-series data sets from heterogeneous sources. Whereas, considering a timeline $T$ and a set of data sources $DS = \left \{DS_1, DS_2, \cdots, DS_m \right \}$  that are used to collect observations related to specific task. We define heterogeneous sources as multiple sources of data that have varying initial time points. Therefore, it cover irregular time periods and generate observations during different time-frame $t\in T$.  Furthermore, for heterogeneous sources, the generated observations have inconsistent features with highly-variable dimensions. 
The goal is to model this heterogeneous temporal data captured from multi-source. This is obtained by designing an architecture that can handle unequal lengths time-series without wasting any useful information, as well as modeling the inconsistent features and manage a huge amount of extracted patterns and correlations.

\subsection{PIETS Architecture }

\name\   is a deep learning architecture that can encode time-series from multi-source data sets. It is designed to overcome the irregular sequence lengths, learn inner data sets patterns and discover correlations between multiple data sets. The full architecture of the proposed model is illustrated in Figure \ref{fig:two}. PIETS consists of three major components: irregularity encoder, parallelised sequence networks and attention mechanism. The overall idea of PIETS can be applied to any deep learning architecture. Also, it is possible to apply PIETS for both multivariate and univariate time-series data sets. 

  \begin{figure}[!t]
  \centering
  \includegraphics[width=\linewidth]{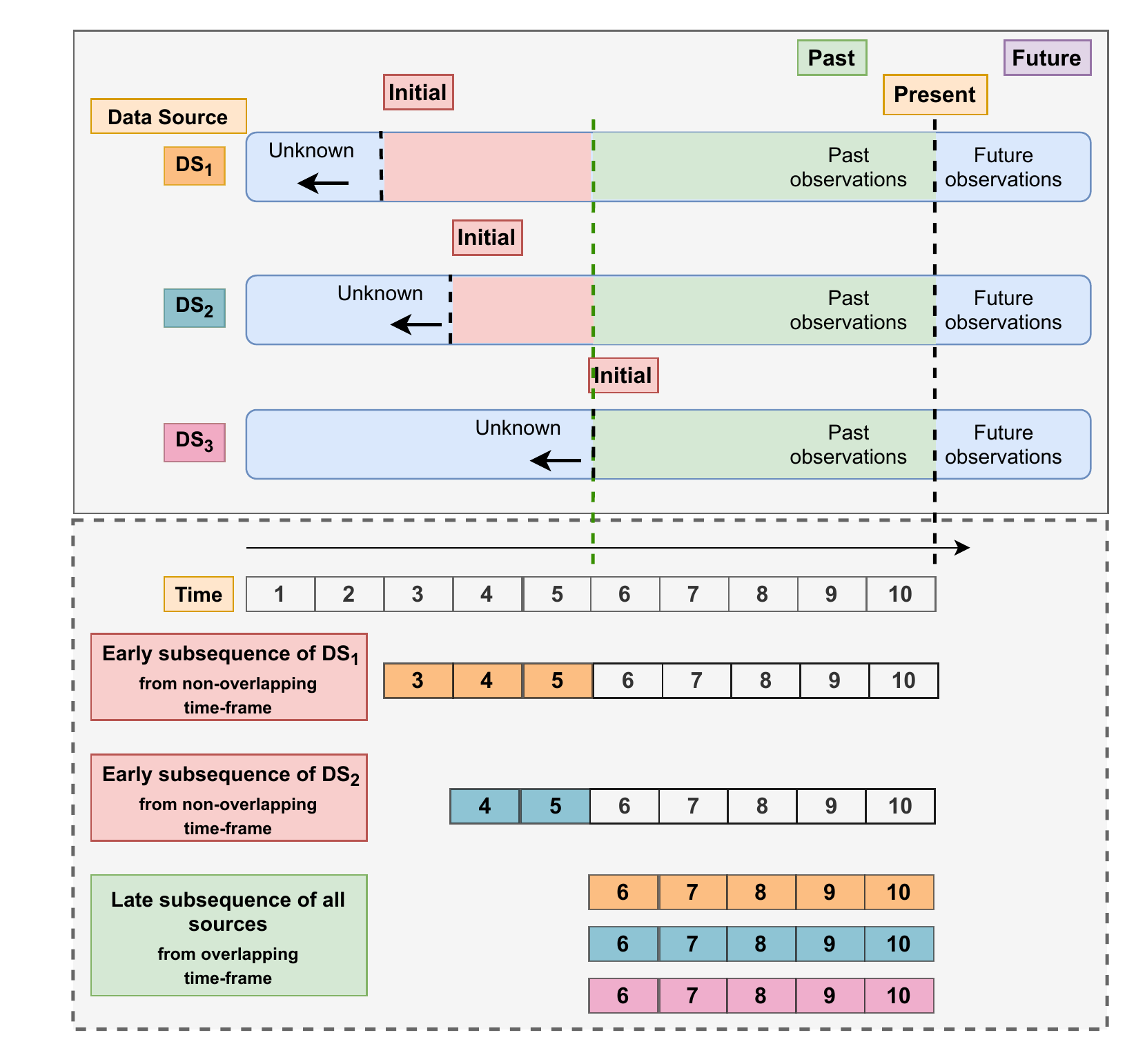}
  \caption{Multi-source time-series snapshot diagram. An example of three multi-source datasets. Each data source has its own initial time point. Red periods identify the sub-sequences of each dataset that considered as non-overlapping sequences. The Green periods represent time series where all sequences have equal size.}
    \label{fig:one}
\end{figure}

\subsubsection{Defining Sub-sequences}
Figure \ref{fig:one} presents a example of three heterogeneous sources of time-series generate temporal observations during irregularity time-frames $t$ belong to  $T = \left \{ 1, 2, .. ,10 \right \}$. As these irregular sequences can not be processed with traditional machine learning models, we overcome this challenge by splitting the sequences of each data source.  As presented in Figure \ref{fig:one}  each data source is remarked with two major time instants (1) The initial point refers to the timestamp where the data source starts recording observation for the first time. (2) Present point which is the current timestamp. Also, each source includes three time periods:  the unknown period where the data set has not started to collect observations yet, the past period which includes the observations made between the initial period and the present point and the future period which will be predicted and observed during the future. By stacking the three data sources together,  two types of sub-sequences are defined. The first one is "Early sub-sequences" which include observations of a Non-overlapping time-frame (marked with red), The non-overlapping time-frame is the time frame where at least one source has not started recording observations, therefore, it includes available sub-sequences of each data set before the latest data source started making observations. These sub-sequences start from the initial point of the data set until the latest initial point available (marked as green dashes in Figure \ref{fig:one}) The second sub-sequence type, is the "Late sub-sequences" (marked with green), which refer to the sub-sequences from the time period where all data are available (overlapping time-frame). Based on this, the non-overlapping sub-sequence of $DS_{1}$ in  Figure \ref{fig:one} include the time points 3,4 and 5, and the non-overlapping sub-sequence of  $DS_{2}$ include points 4 and 5. While time points from 6 to 10 represent an equal size time-series data capture from the three data sources.

 \subsubsection{Irregularity Encoding}
 As mentioned earlier, multi-source time-series imply highly irregular samples. As two sequences from two sources cover different time-frame, it is important to make these sequences equals before processing them with machine learning.  To handle this issue, we proposed an encoding process that does not require imputation or cutting any parts of the data sets, where \name\ encodes the data from each data source separately using two encoding phases based on the previously defined sub-sequences types (early and late sub-sequences). The late sub-sequences that cover the same time frames are used as the primary input for the models, while the dropped early sub-sequences are used to build the learned weights.  As presented in  Figure \ref{fig:two} - Irregularity encoder module, the early sub-sequences from non-overlapped time frames  $\ X_{t} = (T=1,..,m) \in DS_{s}$ are encoded to learn informative hidden representation weights for each data set using any encoding network; $\ LSTM(X_{t})$ in our case. In this phase  \name\  obtain the initialization matrix $\ W_{x_{t}} = (D  \times  N) $. Where $D$ is the dimensions of the sequence and $N$ is the dimensions of the weights space. The first encoding process starts by defining the non-overlapped sub-sequences as Equation (1): 

\begin{equation}
V1 = [X_{1},...,X_{s}]
\end{equation}

Where $s$ is the total number of data sets and $X_{s} =X_{t} =[T= 1,...,m]$. Each $X_{t}$  is passed to sequence model $LSTM(X_{t})$. Then a vector is created for the single data set to include the learned weights $w$ into the form

\begin{equation}
W_{x_{t}} = D \times M
\end{equation}

$D$ is the number of features observed by the data set and $N$ is the dimensions of the weights. In the second phase, all the overlapped sub-sequences from all data sets are defined and encoded as Equation (3):

\begin{equation}
V2 = [X_{1},...,X_{s}]
\end{equation}

Where $s$ is the total number of data sets $\ X_{s} =X_{t} =[T= 1,...,T]$. Each $X_{t}$  is passed to sequence model along with the learned weights matrix  $\ W_{x_{t}} $. Therefore the sequence model $(f)$ takes the data set as an input and the corresponding weights as an initializer 

\begin{equation}
f = LSTM(X_{t} , W_{x_{t}})
\end{equation}

Injecting weights to the model have many preferences over other fusion and joining data methods. By splitting the sequences into two types, we address the irregularity problem associated with having multi-source data since the last features that will be joined together have unified ranges. Unlike using fixed-size windows and modeling only specific parts of the data, our model can use all information to emphasize the encoding and modeling process. Moreover, by addressing only the weights of the data (rather than the raw data) the model reduces the chances of data overwhelming and over-fitting.

\subsubsection{Parallelised Time-Series Modeling}
As previously described, sequence models are used to learn and encode the features. After having the initialization weights from non-overlapping sequences, each data set is modeled separately. Parallel deep learning networks are used and initialized by the corresponding weights matrix $\ W_{x_{t}} $.  Any type of neural network could be applied. For our experiments, we used LSTM and Dense layers with the Rectified Linear Unit (ReLU) as inner activation for hidden layers, ReLU ($X \times W + B$).

As shown in Figure \ref{fig:two}, the main data is fed into two models sequentially. The first one is the parallelized networks, where the inner relations and temporal patterns between the data set observations are captured. The second model is unified hidden layers that discover and learn the patterns between information from different data sets. Toward this, a concatenate layer is used to combine the output of all parallelized networks and a sequential model is then used to catch any additional temporal correlation.

Since our model is designed for multi-source data, having a parallel neural network preserves high flexible model as several types of neural networks could be used based on the type and structure of the raw data. Furthermore, it helps learn and extract the most related features to a specific task. So as the goal of the parallel layers is to independently learn the correlation between features from the same source, having an additional unified network for combined data enhances the extracted features and learns any hidden correlation between data sets.

\subsubsection {Attention Mechanism }
We add an attention layer before the final output to help the model focus on the different periods of the sequences and give different importance for different sequences. The goal is to learn the importance of individual sequences by highlighting specific parts of each data sample with an importance score. In this step, all hidden states for each sequence are passed to the attention network with input shape of  (batch-size, time-steps, input-dimension). Where first, a score function ($ score(h_{t}, h_{s})$) is trying to compute an alignment score given a target hidden state ($h_t$) and source hidden states ($h_s$).

 \begin{equation}
 score (h_{t},\overline{h_{s}})= h_{t}^{T}\cdot  W_{a} \cdot h_{s}
 \end{equation}
 
here $W$ is the trainable weights matrix of attention. After having the attention score, the attention weights $ \alpha _{ts} $ for each sequence is calculated using softmax as follow:

\begin{equation}
\alpha _{ts} = \frac{exp (score (h_{t},\overline{h_{s}}))}{\sum_{{s}'=1}^{S}exp (score (h_{t},\overline{h_{s'}}))}
\end{equation}

Finally, the attention vector is passed as input to a fully connected layer based on context vector  $c_{t}$, where:

\begin{equation}
c_{t} = \sum_{s} \alpha _{ts}\overline{h_{s}}
\end{equation}

\begin{equation}
\alpha _{t} = f(c_{t}, h_{t}) = tanh(W_{c}[c_{t};h_{t}])
\end{equation}

\section{Experiments}

\subsection{COVID-19 Data Sets}
We apply our method to COVID-19 forecasting task. Forecasting COVID-19 is an extremely hard task where the outbreak of the virus depends on many factors related to the nature of the virus and the applied safety procedures. Therefore it is required information from many sectors to be able to analyze and predict the outbreak of COVID-19.  Accordingly,  multi-source time-series data related to COVID-19 have been collected and used to test the performance of the proposed model. These data sets highly demonstrate our problem statement, as each data source has collected information for different periods of time. We used seven data sets related to five basic domains including COVID-19 historical information, geographical, health, mobility and people's behavior data.

\subsubsection{Historical data}
For the historical COVID-19 information, we extracted the daily number of confirmed new cases per state from Wikipedia: COVID-19 pandemic data - United states medical cases.
\subsubsection{Geographical data}
Geographical information is represented by the daily number of population and density of people extracted from the aggregated population data set from Facebook Disaster Maps \cite{maas2019facebook}.
\subsubsection{Health information}
Health information was collected from Aggregated COVID-19 Symptom Survey data set \cite{kreuter2020partnering}. The data was obtained from  Facebook users and it includes self-report COVID-19 related symptoms.
\subsubsection{Mobility information}
We extracted mobility data from three data sets: Google Community Mobility Reports data set., Apple Mobility Trend Reports and Movement data from Facebook Disaster Maps \cite{maas2019facebook}.
\subsubsection{Behavior information}
Behavior data is represented by Google search trends, We used the search trends data for COVID-19 symptoms from Google Cloud Platform-COVID-19-open-data and we manually extracted additional data from Google trend related to the following search terms (COVID-19–Coronavirus, How to know if I have COVID-19, How to know if I have coronavirus, How do I know if I have COVID-19, How do I know if I have coronavirus, What are COVID-19 symptoms, What are coronavirus symptoms). 

\subsection{Setup and Hyper-parameter Settings}
\subsubsection {Sliding Windows}
To construct the sequences from collected data sets we used a sliding window of seven days length, which means that the model predicts the outbreak of day $t$ after studying and learning the features of the previous days from $t-7$ to $t-1$. Each data sample is represented with a set of sequences and the features vector is $M \times D$, where $M$ is the number of sequences (number of time points), seven in our case, and $D$ is the number of features that observed by specific data-source.

We determine the lag size based on the auto-correlation of the time-series and the virus incubation period. We study the auto-correlation of COVID-19 daily confirmed cases using the Auto-Correlation Functions (ACF) and Partial-ACF. Through the analysis of the data set, we found that ACF and PACF demonstrate a high correlation between the daily confirmed number of cases and its latest ten recent values. Also, ACF showed slow decay which implies that the future values have a very high correlation with their past values. However, since the average incubation time for COVID-19 is 4 to 7 days, we choose the lag size to be 7.


\subsubsection {Hyper-parameter}
For our experiments, five different US states have been used  (New York (NY), Illinois (IL), North Carolina (NC), Wisconsin (WI), Oklahoma (OK)) , these selected states have various levels of COVID-19 outbreak. Each data set and state represents and covers different time intervals.  For all experiments, non-shuffled data splitting into a training and testing set have been applied. 66\% of the sequences were used for training while the testing set held the remaining 33\%. We used 20\% of the training data for validation. Only the sequences marked as overlapping sequences were split, while for the irregularity encoder all early sequences were used for training. LSTM model was used as our main model with an output dimension size equal to 128. Parallel LSTM layers have been used to learn the intra-dataset correlations and patterns, followed by a concatenation layer and another LSTM network to learn the datasets correlation. We run each experiment including the baseline 5 times using several random seeds to initialize the model parameters. The model is trained with Adam optimizer and implemented using Tensorflow on a desktop with an NVIDIA GeForce MX230.

\subsection{Performance Evaluation}
\subsubsection{Evaluation Protocol}
As described in the problem statement, the data sets are irregular over long time intervals thus include long continuous sequences of missing observations. Since this issue has not been explicitly proposed before, we explore two potential techniques as baselines to handle this issue. We conduct a wide range of experiments using features level fusion along with regular neural network and irregular time series modeling. Features level fusion is one of the most common fusion methods for dealing with multi-source data. In order to have features level fusion, all sequences and time-series data are required to have the same length to be fed into the machine learning model. Therefore, a predefined fixed-size window is needed, which is a primitive way for handling unequal size data samples by performing a fixed size window $n$ to all data sequences. The windows cover sub-sequences of the series where all observations are available. Each sequence has a length equal to $n$ and only the selected parts of the data sets are used for training and testing while the reminding parts are discarded.

For the irregular models, we evaluate the most recent methods that have been designed for irregular time series data (decay and ODE-based models). These methods have shown a good ability to handle missing values and are used to handle data sets with different interval sizes between observations (e.g.unequal and irregular observed samples). Overall, the following models are considered for comparison:


\begin{itemize}
\item GRU-D: Gated Recurrent Unit (GRU) based model with decay method that is performed on two-part; decay for input to perform input imputation and exponential decay for the hidden state. The model captures the informative missingness by incorporating masking and time interval. It required three inputs including the original sequence (observed variable),  masking vector to inform if the input is observed or missing and time interval to represents the time patterns of observations \cite{che2018recurrent}.
\item RNN-VAE: A variational autoencoder consists of encoder and decoder neural network with standard RNN model for each.
\item ODE-RNN: RNN neural network with a continuous hidden state that is calculated based on neural ODEs and standard  RNN. Whereas neural ODEs are applied to model hidden state dynamics when there are no observations, standard RNN is used to update the hidden state when a new observation appears \cite{rubanova2019latent}.
\item Latent ODE: Encoder-decoder architectures with ODE-RNN network for encoder and neural-ODE for decoder \cite{rubanova2019latent}.
\item Neural CDE:  an extended Neural- ODE model using controlled differential equations to increase the ability for the model to processing incoming data and adjust the variable trajectories. Neural-CDE trained with memory-efficient adjoint-based backpropagation even across observations. \cite{kidger2020neural}.
\end{itemize}

To assess the performance of our model, we use Root Mean Squared Error (RMSE), Mean Squared Error (MSE), and Mean Absolute Percentage Error (MAPE).

\subsubsection{Evaluation Result}

Table~\ref{tab:table1} presents the results of our model against several baselines. The first method describes the result using a predefined window and feature-level fusion. Where specific parts of each data set are selected and all features combined as one vector of equal-size sequences. The generated data is fed to an LSTM model. As the result indicates, using  \name\ achieves a higher performance in terms of MSE, RMSE comparing with using the primitive data fusion and one network.
The rest of the table presents the results using the irregularity specialist models and their performance to handle the heterogeneous multi-source time-series. Stander RNN along with Autoencoders continued to perform poorly for all states. Also, Neural CDE showed the same poor performance. On the other hand, ODE-RNN reported the best performance among all baselines especially for New York and North Carolina that have the less percentage of missing observations. The rest of the baseline models (latent-ode and GRU-D) have not shown a great performance yet they performed much better comparing with CDE and RNN-VAE models. The final result shows that our \name\ model outperforms these state-of-the-art models where the irregularity model was unable to handle the long sequences of missing observation accrue with multi-source data. 

\begin{table*}[!t]
\centering
  \caption{Test MSE, RMSE, MAPE on several COVID-19 related datasets for \name\  and state-of-the-art baseline models}
  \label{tab:table1}
    \begin{tabular}{|l||c||c||c||c||c||c|}
    \hline
    Method & Metrics & New York & Illinois & North Carolina & Oklahoma & Wisconsin \\
    \hline
    {predefined window} & MSE & 0.0036   & 0.4590 & 0.7935 & 0.6439 & 1.3620 \\ & RMSE & 0.0602   & 0.6775 & 0.8898 & 0.7998 & 1.1625 \\ & MAPE & 0.1644 &  0.1878 & 0.3019 & \textbf{0.2143} & 0.2584 \\ \hline
    {RNN-VAE} & MSE & 0.0228 &  1.4203 & 3.1513 & 4.5986 & 4.696 \\
 & RMSE & 0.1482 & 1.1918 & 1.7752 & 2.1444 & 2.167 \\
 & MAPE & 0.5175 & 0.5188 & 0.5187 & 0.7248 & 0.6063 \\ \hline
    {Neural CDE (NeurIPS 2020) \cite{kidger2020neural}} & MSE & 0.7133 & 0.8972 & 1.082 & 3.3308 & 2.8275 \\
 & RMSE & 0.8186  & 0.9472 & 1.0402 & 1.8114 & 1.6764 \\
 & MAPE & 2.1934 &  0.3344 & 0.2533 & 0.5695 & 0.4507 \\ \hline
    {ODE-RNN (arxiv 2019) \cite{rubanova2019latent}} & MSE & 0.0025 & 0.6068 & 0.7252 & 1.1522 & 2.5753 \\
 & RMSE & 0.0495 &  0.7789 & 0.8515 & 1.0734 & 1.6048 \\
 & MAPE & 0.1397  & 0.2937 & 0.3037 & 0.3033 & 0.3378 \\ \hline
    {latent ODE (NeurIPS 2018) \cite{chen2018neural}} & MSE & 0.0031  & 0.7069 & 0.8268 & 1.1563 & 1.4346 \\
 & RMSE & 0.0556 & 0.8408 & 0.9093 & 1.0753 & 1.1977 \\
 & MAPE & 0.1754 &  0.2685 & 0.3402 & 0.3312 & 0.2819 \\ \hline
    {GRU-D (scientific reports 2018) \cite{che2018recurrent}} & MSE & 0.0028  & 0.5851 & 0.8171 & 0.8601 & 3.1178 \\
 & RMSE & 0.0532  & 0.7649 & 0.9039 & 0.9274 & 1.7657 \\
 & MAPE & 0.1502  & 0.2668 & 0.3313 & 0.2818 & 0.4271 \\ \hline
    {\name\ (ours)} & MSE & \textbf{0.0021}  & \textbf{0.3903} & \textbf{0.6488} & \textbf{0.6039} & \textbf{0.844} \\
 & RMSE & \textbf{0.0463} &  \textbf{0.6247} & \textbf{0.8131} & \textbf{0.7814} & \textbf{0.9753} \\
 & MAPE & \textbf{0.1386} &  \textbf{0.1704} & \textbf{0.2292} & 0.219 & \textbf{0.2318} \\
    \hline
  \end{tabular}
\end{table*}

\section{Ablation Study}

In this part, we examine the effect of individual components of our network. Particularly, the effect of weights initializer and attention network. We add the major components to the model one by one and validate their performance using MSE measurement. The following three architectures are considered along with the full architecture of \name:
\begin{itemize}
\item Basic model: Includes only the predictors with random initial weights and a fully connected layer to combine all the features.
\item Attention only model: Includes the same components of the base model along with an attention layer that assigns different importance rates to different features.
\item Weights initializer only model: Includes the same components of the base model with reforms on the initial weights to be defined based on previously learned features. 
\end{itemize}

The results of ablation study are presented in Table~\ref{tab:AblationStudy}.
\begin{table*}
\centering
  \caption{MSE result for ablation study results on several datasets}
    \label{tab:AblationStudy}
     \begin{tabular}{|cc||ccccc|}
        \hline
        \multicolumn{2}{|c||}{Configurations} &  \multicolumn{5}{c|}{Dataset}  \\ \hline
 Weight initilizer & Attention & New York& Illinois & North Carolina & Oklahoma &  Wisconsin \\ \hline
        \hline
        \xmark & \xmark & 0.0025 &0.4903& 0.7149 & 0.8384 & 1.2206 \\
        \cmark & \xmark & 0.0023 & 0.3980 &0.6808& 0.6140 & 0.8634 \\
        \xmark &  \cmark & 0.0032 & 0.3976 &0.6533& 0.6283 & 0.9611 \\
         \cmark &  \cmark & 0.0021 & 0.3900 & 0.6488  & 0.6039 &0.8440 \\
    \hline
  \end{tabular}
\end{table*}
\noindent\textbf{Effect of initial weights:} As shown in the second row of Table~\ref{tab:AblationStudy}, the weights initializer has a better performance than the base model (the first row). By only introducing the initial weights, on each dataset, the MSE value has dropped by 8\%, 18.83\%, 4.77\%, 26.77\%, and 29.26\%, respectively compared to the base model.
 
\noindent\textbf{Effect of attention:} The performance of only adding the attention component is presented in the third row of Table~\ref{tab:AblationStudy}. Although it does not improve the performance as much as the weights initializer module, it still outperforms the base model for most datasets except for New York. On average, using the attention module the MSE values decreased by 18.46\% over the base model.

In general, the configuration with both weight initilizer and attention mechanism enabled outperforms all other three configurations. This result justifies the need of incorporating these two components in \name.

\section{Model Interpretability and Robustness}
\begin{figure*}[!t]
\centering
\subcaptionbox{New York\label{Attention_NewYork}}
{\includegraphics[width=0.25\textwidth]{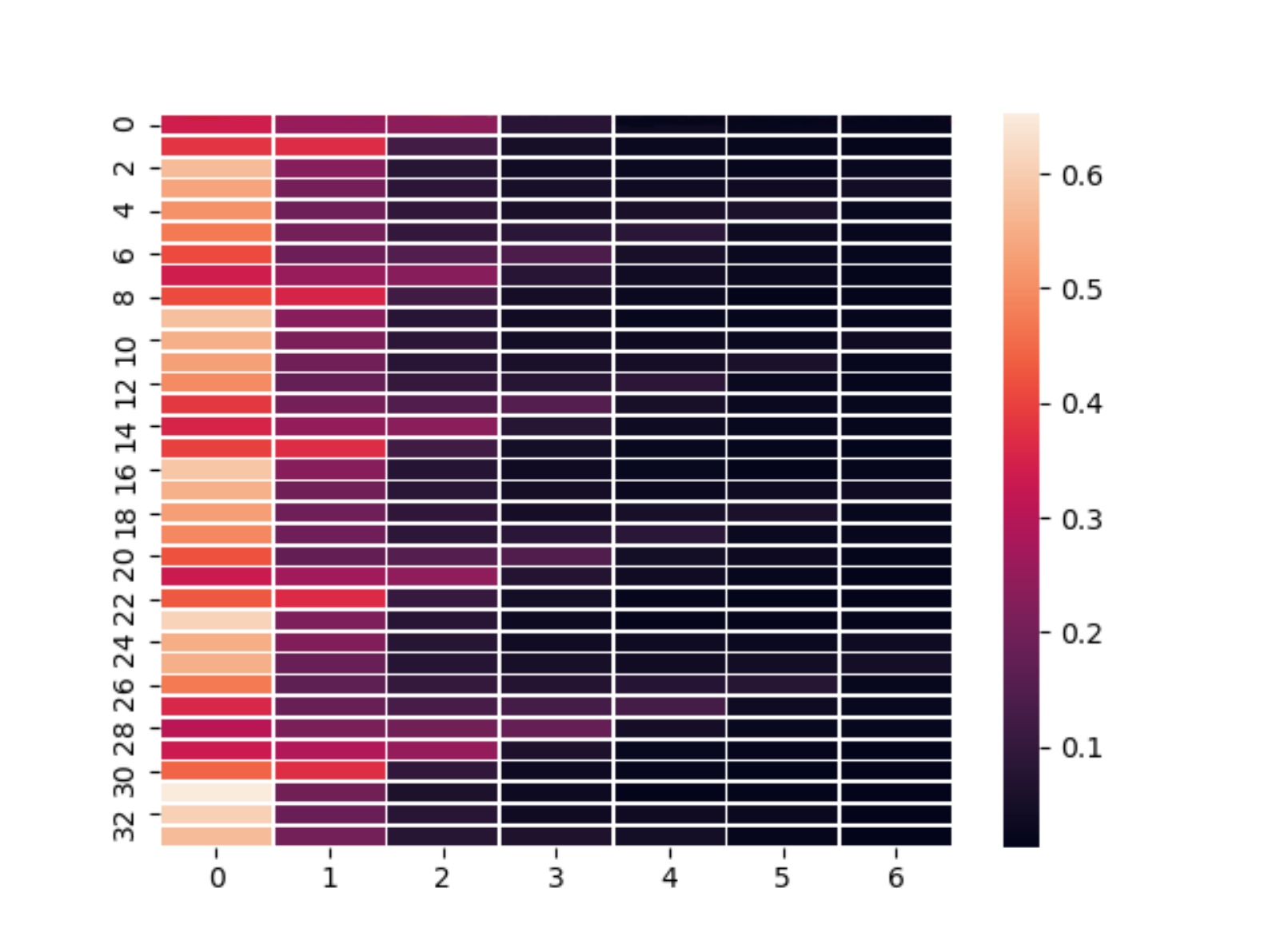}}
\subcaptionbox{North Carolina\label{Attention_NorthCarolina}}
{\includegraphics[width=0.25\textwidth]{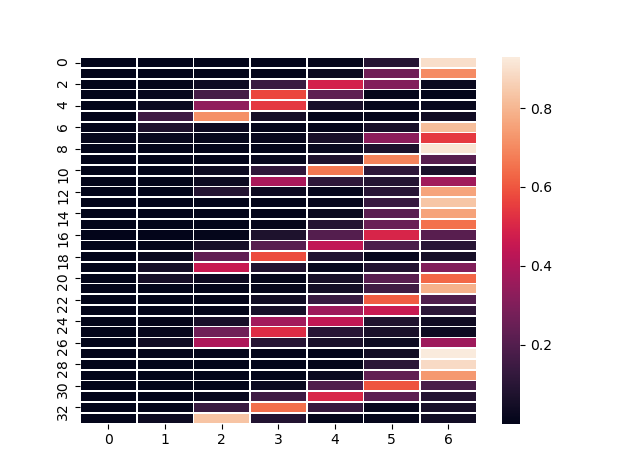}}
\subcaptionbox{Illinois\label{Attention_Illinois}}
{\includegraphics[width=0.25\textwidth]{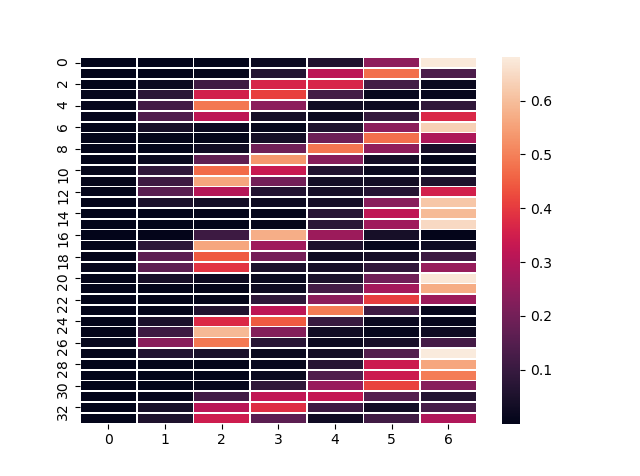}}
\subcaptionbox{Oklahoma\label{Attention_Oklahoma}}
{\includegraphics[width=0.25\textwidth]{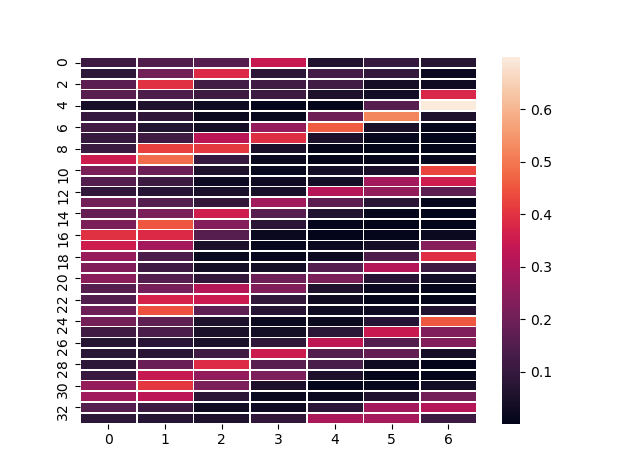}}
\subcaptionbox{Wisconsin\label{Attention_Wisconsin}}
{\includegraphics[width=0.25\textwidth]{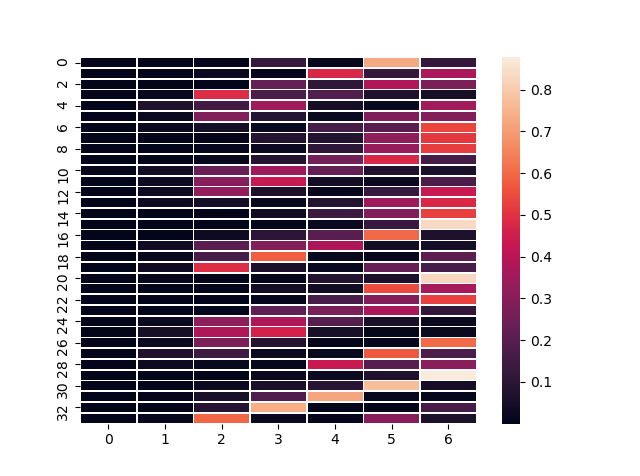}}
\caption{Illustration of model interpreterability in terms of the attention weights for all case studies. }
\label{fig:Attention}
\end{figure*}

To explore the robustness of our method we analyze loss convergence and attention importance weights in the section. From Figure~\ref{fig:Attention} we can see that the model gives different attention weights for different sequences of lag. This indicates that the model learns the importance of each sequence and its correlation to the current target day. Therefore, prevents irrelevant observations from having a huge influence on the result. For most of the cases, the oldest lags get the highest importance rates which are sensible as mostly the outbreak of COVID-19 relies more on the movement and behaviors of people in the last 3 to 7 days rather than the behaviors in the last one day. For example, in Figure \ref{fig:Attention}(b),(c) and (e) a very low importance rate has been given to first and second lag sequences. On the other hand for New York and Oklahoma Figure \ref{fig:Attention}(a) and (d) the attention highlighted the early lags sequences.

\begin{figure*}[!t]
\centering
\subcaptionbox{New York\label{LOSS_NewYork}}
{\includegraphics[width=0.30\textwidth]{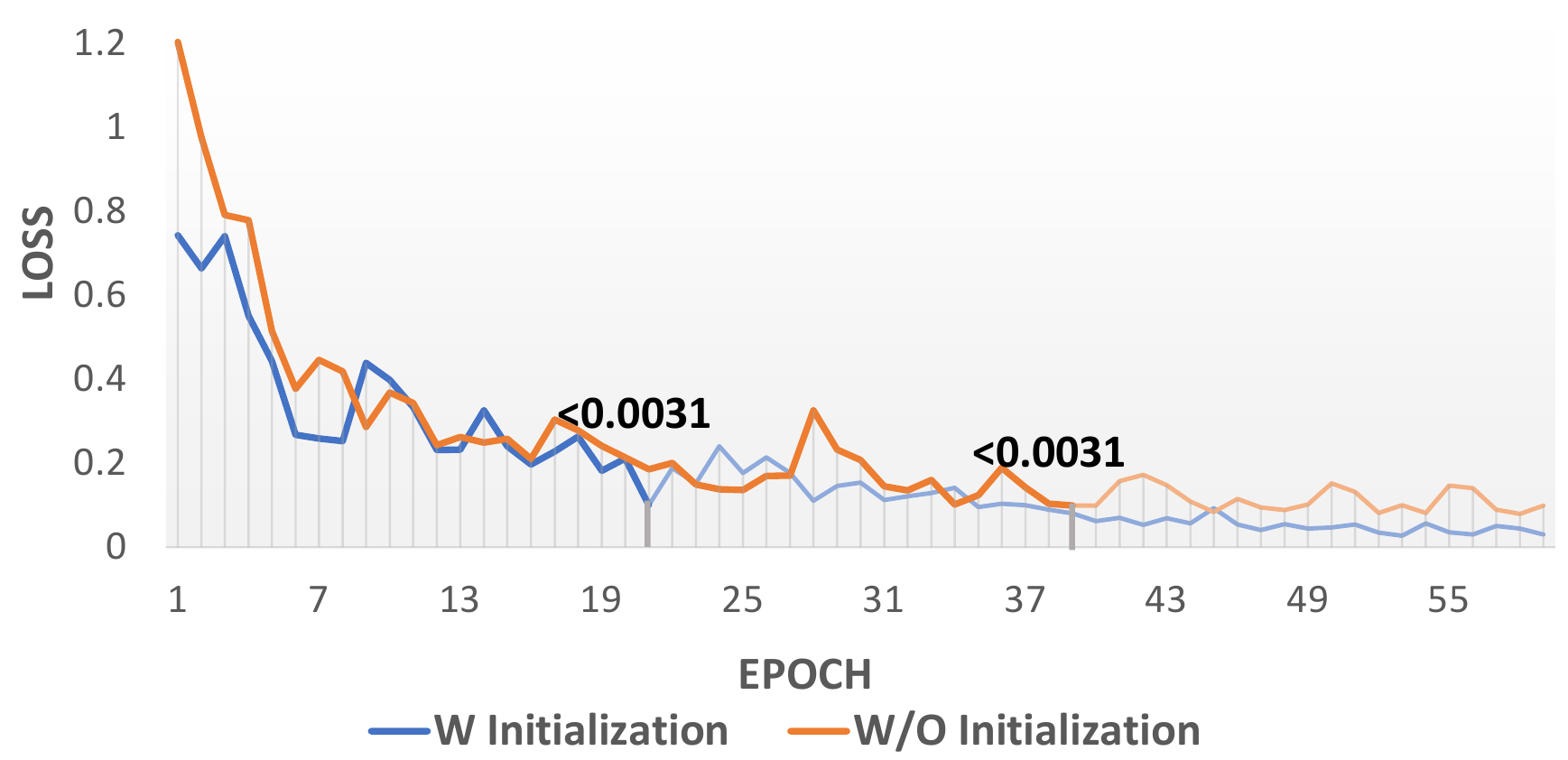}}
\subcaptionbox{North Carolina\label{LOSS_NorthCarolina}}
{\includegraphics[width=0.30\textwidth]{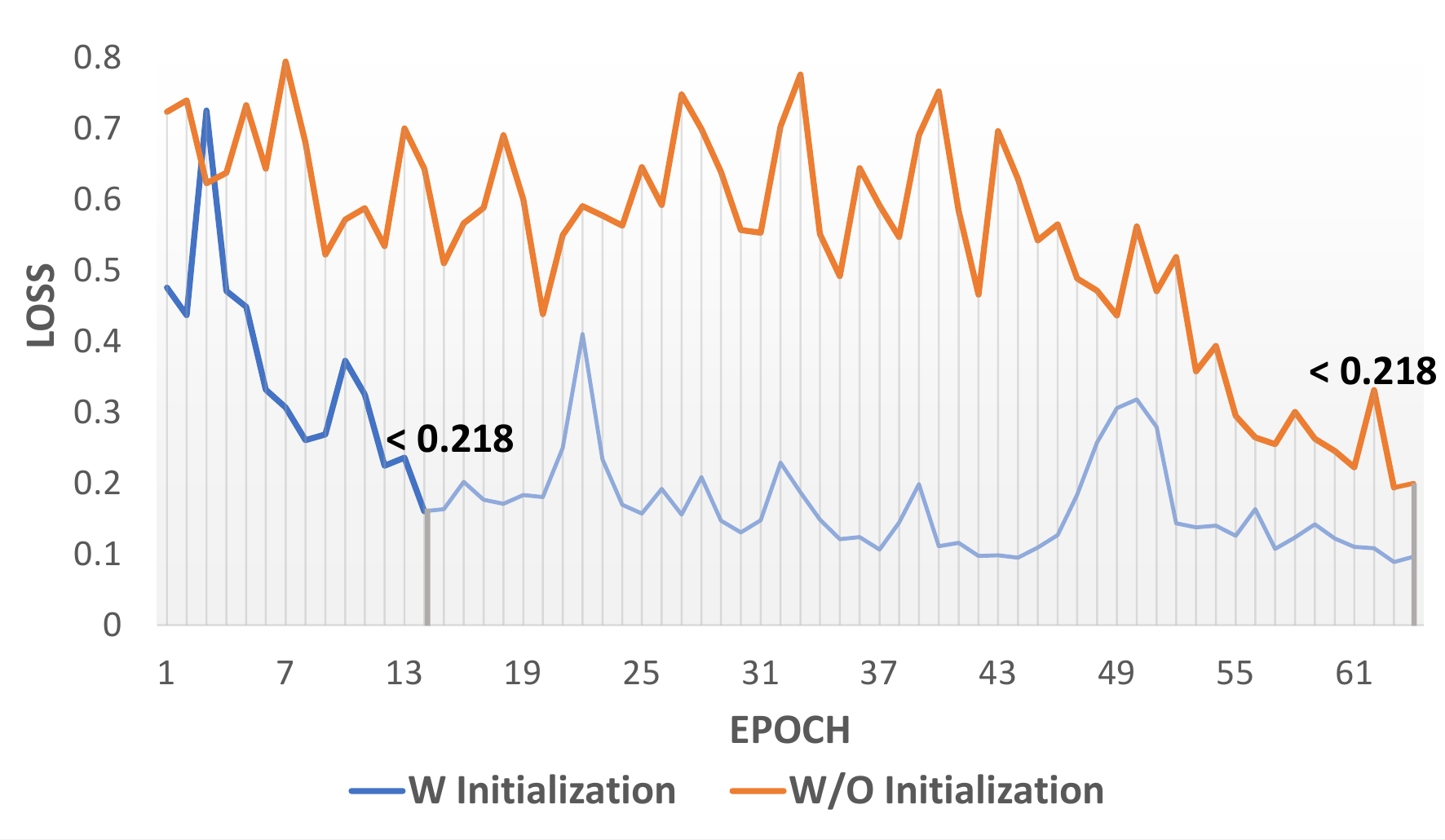}}
\subcaptionbox{Illinois\label{LOSS_Illinois}}
{\includegraphics[width=0.30\textwidth]{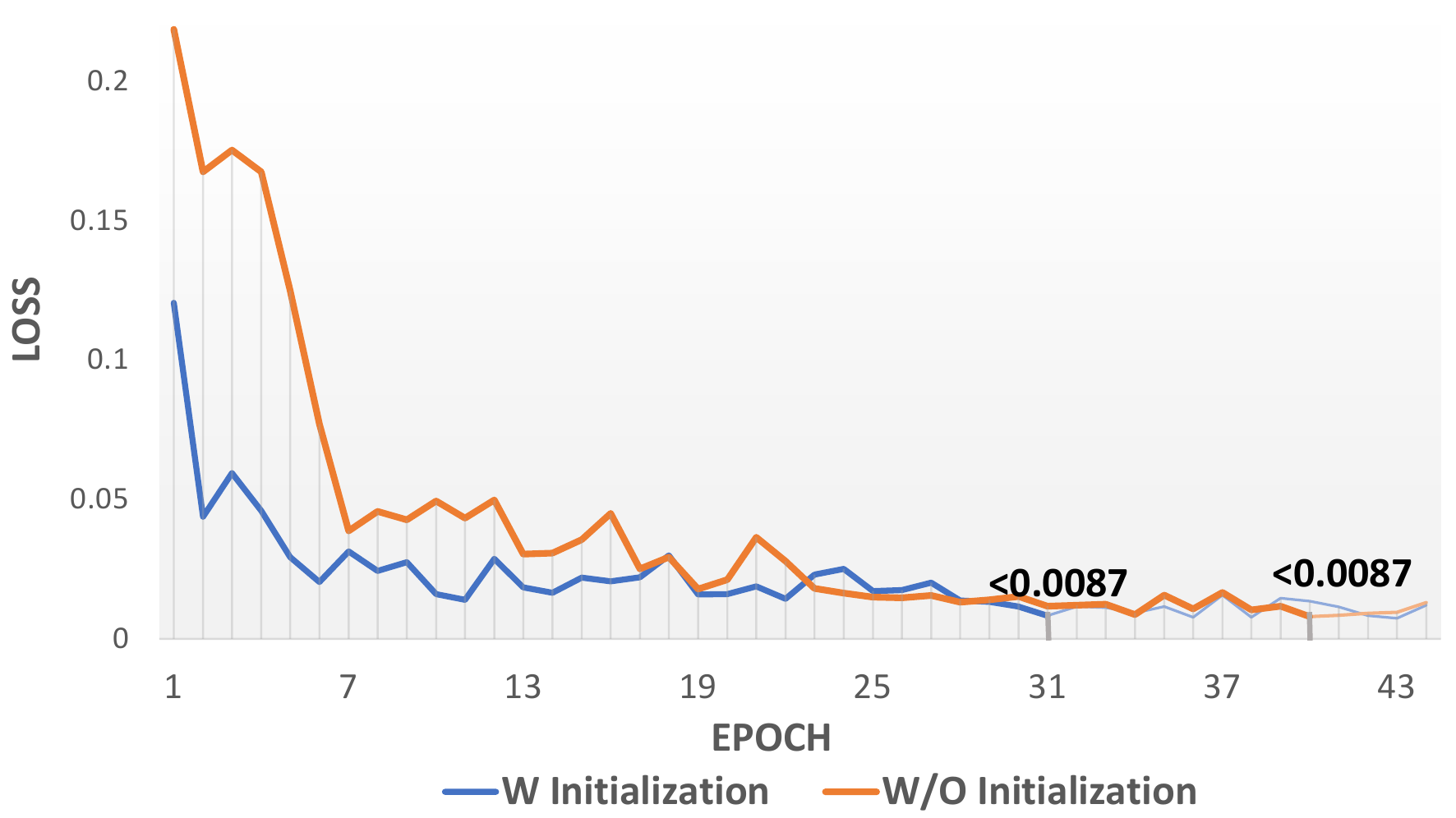}}
\subcaptionbox{Oklahoma\label{LOSS_Oklahoma}}
{\includegraphics[width=0.30\textwidth]{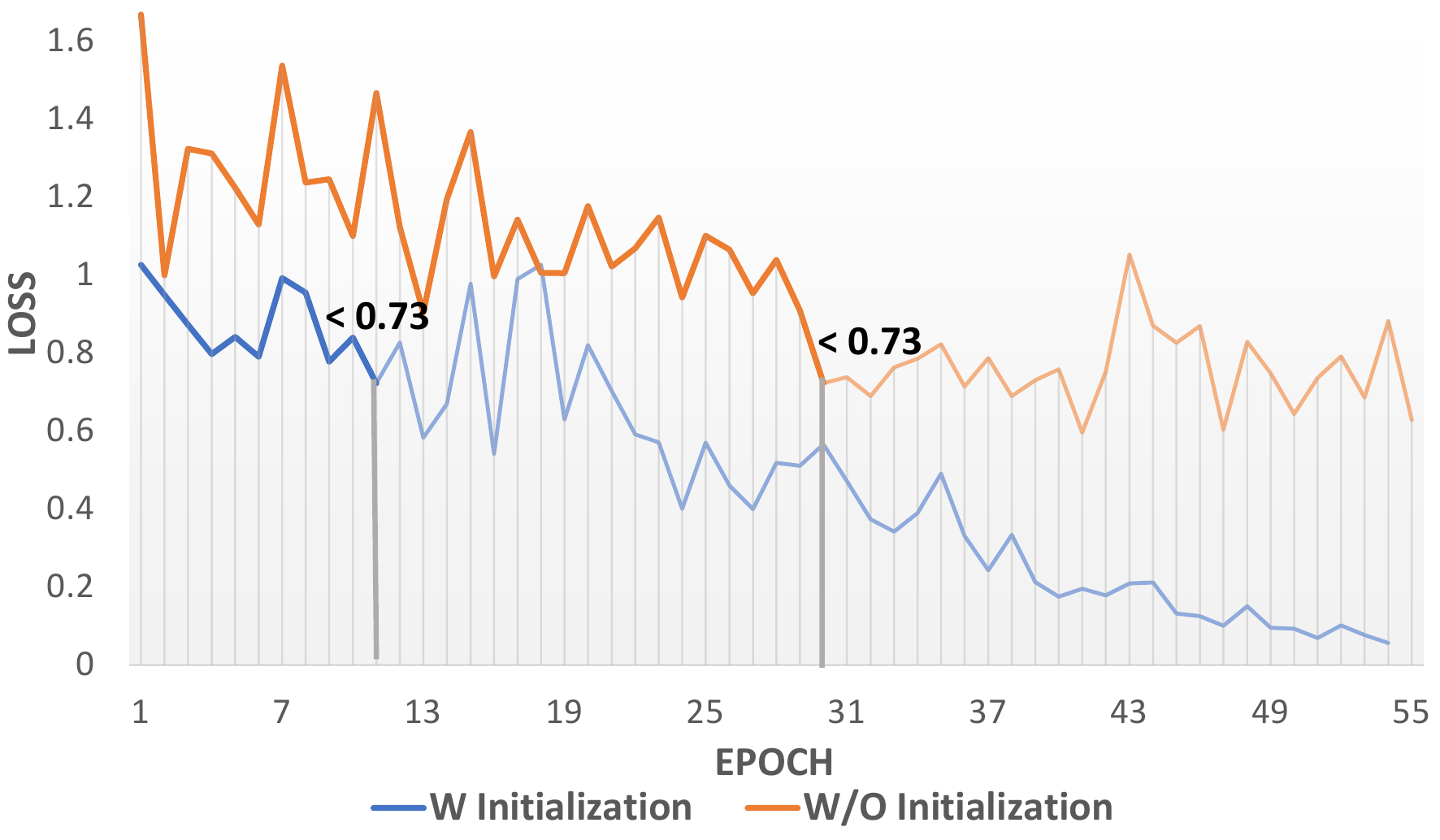}}
\subcaptionbox{Wisconsin\label{LOSS_Wisconsin}}
{\includegraphics[width=0.30\textwidth]{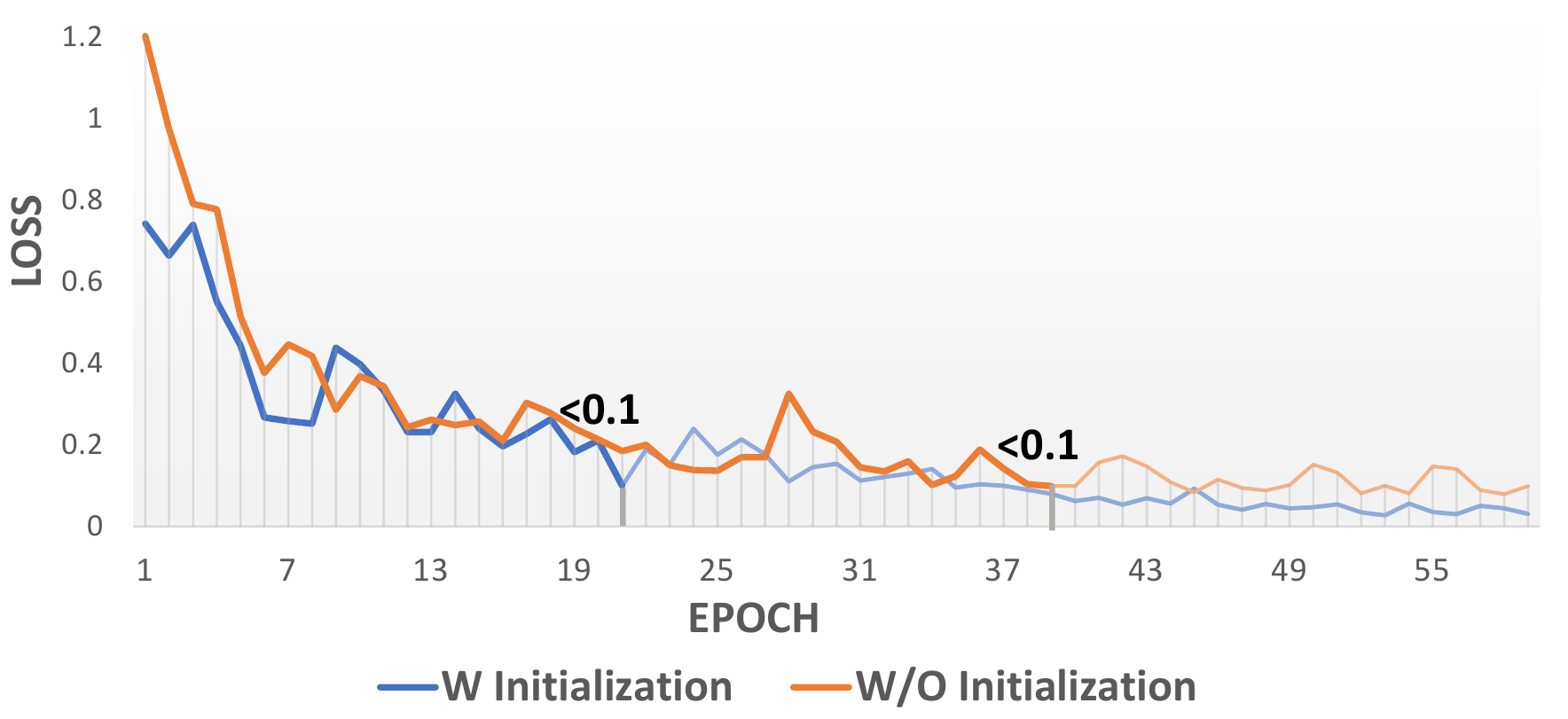}}
\caption{Loss convergence during training for \name\ with and without weights initializer. Using the full modules of \name\  accelerates the convergence for all cases. }\label{fig:LOSS}
\end{figure*}

Figure~\ref{fig:LOSS} presents the model training loss with and without the weights initializer for each data set. We show that using the pre-trained weights extracted from the irregularity encoder leads to a faster converging of the neural network during the training period. Considering North Carolina as an example (Figure \ref{fig:LOSS}(b)), without adding the weights initializer, the model requires an extra 52 epochs to achieve a loss value less than the average loss values achieved by the model at the first 100 epochs (0.218). This also applies to New York, Oklahoma and Wisconsin (Figure~\ref{fig:LOSS}(a),(d) and (e)), where the model requires many extra epochs (19 extra epochs for Oklahoma and 18 extra epochs for New York and Wisconsin) to reach the average loss values achieved at the first 100 epochs. While for (Illinois, Figure~\ref{fig:LOSS}(c)), the weights initializer has a relatively less effect on convergence, as using the weights initializer accelerated the convergence by 8 epochs.

\section{conclusion}
In this paper, we present a state-of-the-art deep learning architecture (\name) for forecasting time series from heterogeneous data sources. Our architecture is designed with several components to model the unequal lengths data sets and effectively learn the patterns and correlation between observations from multi-source. \name\ yields a competitive performance on the challenging COVID-19 case number forecasting task with real-world data sets. Unlike other traditional fusion methods, \name\ does not rely on imputing or discarding any data samples, which results in a better ability to enhance the prediction.
Furthermore, our experiments demonstrate that using pre-trained weights generated from \name\ irregularity encoder accelerates the convergence of the model and improves the forecasting performance. Moreover, we show that using an attention mechanism increases the interpretability of the model and highlights the most related feature sets.

\section*{Acknowledgment}
 This research is supported by Australian Research Council (ARC) Discovery Project \textit{DP190101485}.
 
\bibliographystyle{IEEEtran}
\bibliography{main}

\begin{thebibliography}{10}
\providecommand{\url}[1]{#1}
\csname url@samestyle\endcsname
\providecommand{\newblock}{\relax}
\providecommand{\bibinfo}[2]{#2}
\providecommand{\BIBentrySTDinterwordspacing}{\spaceskip=0pt\relax}
\providecommand{\BIBentryALTinterwordstretchfactor}{4}
\providecommand{\BIBentryALTinterwordspacing}{\spaceskip=\fontdimen2\font plus
\BIBentryALTinterwordstretchfactor\fontdimen3\font minus
  \fontdimen4\font\relax}
\providecommand{\BIBforeignlanguage}[2]{{%
\expandafter\ifx\csname l@#1\endcsname\relax
\typeout{** WARNING: IEEEtran.bst: No hyphenation pattern has been}%
\typeout{** loaded for the language `#1'. Using the pattern for}%
\typeout{** the default language instead.}%
\else
\language=\csname l@#1\endcsname
\fi
#2}}
\providecommand{\BIBdecl}{\relax}
\BIBdecl

\bibitem{horn2020set}
M.~Horn, M.~Moor, C.~Bock, B.~Rieck, and K.~M. Borgwardt, ``Set functions for
  time series,'' in \emph{Proceedings of the 37th International Conference on
  Machine Learning, {ICML} 2020, 13-18 July 2020, Virtual Event}, ser.
  Proceedings of Machine Learning Research, vol. 119.\hskip 1em plus 0.5em
  minus 0.4em\relax {PMLR}, 2020, pp. 4353--4363.

\bibitem{martinez2015survey}
F.~Mart{\'\i}nez-{\'A}lvarez, A.~Troncoso, G.~Asencio-Cort{\'e}s, and J.~C.
  Riquelme, ``A survey on data mining techniques applied to electricity-related
  time series forecasting,'' \emph{Energies}, vol.~8, no.~11, pp.
  13\,162--13\,193, 2015.

\bibitem{lim2020time}
B.~Lim and S.~Zohren, ``Time-series forecasting with deep learning: a survey,''
  \emph{Philosophical Transactions of the Royal Society A}, vol. 379, no. 2194,
  p. 20200209, 2021.

\bibitem{mahalle2020data}
P.~N. Mahalle, N.~P. Sable, N.~P. Mahalle, and G.~R. Shinde, ``Data analytics:
  Covid-19 prediction using multimodal data,'' in \emph{Intelligent Systems and
  Methods to Combat Covid-19}.\hskip 1em plus 0.5em minus 0.4em\relax Springer,
  2020, pp. 1--10.

\bibitem{DBLP:journals/sncs/ShindeKMDCH20}
G.~R. Shinde, A.~B. Kalamkar, P.~N. Mahalle, N.~Dey, J.~Chaki, and A.~E.
  Hassanien, ``Forecasting models for coronavirus disease {(COVID-19):} {A}
  survey of the state-of-the-art,'' \emph{{SN} Comput. Sci.}, vol.~1, no.~4, p.
  197, 2020.

\bibitem{DBLP:journals/jms/Santosh20}
K.~C. Santosh, ``Ai-driven tools for coronavirus outbreak: Need of active
  learning and cross-population train/test models on multitudinal/multimodal
  data,'' \emph{J. Medical Syst.}, vol.~44, no.~5, p.~93, 2020.

\bibitem{shukla2021multi}
S.~N. Shukla and B.~M. Marlin, ``Multi-time attention networks for irregularly
  sampled time series,'' \emph{arXiv preprint arXiv:2101.10318}, 2021.

\bibitem{sherstinsky2020fundamentals}
A.~Sherstinsky, ``Fundamentals of recurrent neural network (rnn) and long
  short-term memory (lstm) network,'' \emph{Physica D: Nonlinear Phenomena},
  vol. 404, p. 132306, 2020.

\bibitem{pratap2019multi}
B.~Pratap~Singh, I.~Deznabi, B.~Narasimhan, B.~Kucharski, R.~Uppaal,
  A.~Josyula, and M.~Fiterau, ``Multi-resolution networks for flexible
  irregular time series modeling (multi-fit),'' \emph{arXiv e-prints}, pp.
  arXiv--1905, 2019.

\bibitem{che2018recurrent}
Z.~Che, S.~Purushotham, K.~Cho, D.~Sontag, and Y.~Liu, ``Recurrent neural
  networks for multivariate time series with missing values,'' \emph{Scientific
  reports}, vol.~8, no.~1, pp. 1--12, 2018.

\bibitem{kim2018temporal}
Y.-J. Kim and M.~Chi, ``Temporal belief memory: Imputing missing data during
  rnn training.'' in \emph{In Proceedings of the 27th International Joint
  Conference on Artificial Intelligence (IJCAI-2018)}, 2018.

\bibitem{zhang2021feature}
C.~Zhang, H.~Fanaee-T, and M.~Thoresen, ``Feature extraction from unequal
  length heterogeneous ehr time series via dynamic time warping and tensor
  decomposition,'' \emph{Data Mining and Knowledge Discovery}, pp. 1--25, 2021.

\bibitem{de2019gru}
E.~D. Brouwer, J.~Simm, A.~Arany, and Y.~Moreau, ``Gru-ode-bayes: Continuous
  modeling of sporadically-observed time series,'' in \emph{Advances in Neural
  Information Processing Systems 32: Annual Conference on Neural Information
  Processing Systems 2019, NeurIPS 2019, December 8-14, 2019, Vancouver, BC,
  Canada}, H.~M. Wallach, H.~Larochelle, A.~Beygelzimer,
  F.~d'Alch{\'{e}}{-}Buc, E.~B. Fox, and R.~Garnett, Eds., 2019, pp.
  7377--7388.

\bibitem{rubanova2019latent}
Y.~Rubanova, R.~T. Chen, and D.~Duvenaud, ``Latent odes for irregularly-sampled
  time series,'' \emph{arXiv preprint arXiv:1907.03907}, 2019.

\bibitem{cao2018brits}
W.~Cao, D.~Wang, J.~Li, H.~Zhou, L.~Li, and Y.~Li, ``{BRITS:} bidirectional
  recurrent imputation for time series,'' in \emph{Advances in Neural
  Information Processing Systems 31: Annual Conference on Neural Information
  Processing Systems 2018, NeurIPS 2018, December 3-8, 2018, Montr{\'{e}}al,
  Canada}, S.~Bengio, H.~M. Wallach, H.~Larochelle, K.~Grauman,
  N.~Cesa{-}Bianchi, and R.~Garnett, Eds., 2018, pp. 6776--6786.

\bibitem{liu2019naomi}
Y.~Liu, R.~Yu, S.~Zheng, E.~Zhan, and Y.~Yue, ``{NAOMI:} non-autoregressive
  multiresolution sequence imputation,'' in \emph{Advances in Neural
  Information Processing Systems 32: Annual Conference on Neural Information
  Processing Systems 2019, NeurIPS 2019, December 8-14, 2019, Vancouver, BC,
  Canada}, H.~M. Wallach, H.~Larochelle, A.~Beygelzimer,
  F.~d'Alch{\'{e}}{-}Buc, E.~B. Fox, and R.~Garnett, Eds., 2019, pp.
  11\,236--11\,246.

\bibitem{ma2020midia}
Q.~Ma, W.~Lee, T.~Fu, Y.~Gu, and G.~Yu, ``{MIDIA:} exploring denoising
  autoencoders for missing data imputation,'' \emph{Data Min. Knowl. Discov.},
  vol.~34, no.~6, pp. 1859--1897, 2020.

\bibitem{khayati2020mind}
M.~Khayati, A.~Lerner, Z.~Tymchenko, and P.~Cudr{\'{e}}{-}Mauroux, ``Mind the
  gap: An experimental evaluation of imputation of missing values techniques in
  time series,'' \emph{Proc. {VLDB} Endow.}, vol.~13, no.~5, pp. 768--782,
  2020.

\bibitem{li2016scalable}
S.~C. Li and B.~M. Marlin, ``A scalable end-to-end gaussian process adapter for
  irregularly sampled time series classification,'' in \emph{Advances in Neural
  Information Processing Systems 29: Annual Conference on Neural Information
  Processing Systems 2016, December 5-10, 2016, Barcelona, Spain}, D.~D. Lee,
  M.~Sugiyama, U.~von Luxburg, I.~Guyon, and R.~Garnett, Eds., 2016, pp.
  1804--1812.

\bibitem{kidger2020neural}
P.~Kidger, J.~Morrill, J.~Foster, and T.~J. Lyons, ``Neural controlled
  differential equations for irregular time series,'' in \emph{Advances in
  Neural Information Processing Systems 33: Annual Conference on Neural
  Information Processing Systems 2020, NeurIPS 2020, December 6-12, 2020,
  virtual}, H.~Larochelle, M.~Ranzato, R.~Hadsell, M.~Balcan, and H.~Lin, Eds.,
  2020.

\bibitem{lipton2016directly}
Z.~C. Lipton, D.~C. Kale, and R.~C. Wetzel, ``Directly modeling missing data in
  sequences with rnns: Improved classification of clinical time series,'' in
  \emph{Proceedings of the 1st Machine Learning in Health Care, {MLHC} 2016,
  Los Angeles, CA, USA, August 19-20, 2016}, ser. {JMLR} Workshop and
  Conference Proceedings, F.~Doshi{-}Velez, J.~Fackler, D.~C. Kale, B.~C.
  Wallace, and J.~Wiens, Eds., vol.~56.\hskip 1em plus 0.5em minus 0.4em\relax
  JMLR.org, 2016, pp. 253--270.

\bibitem{yadav2018mining}
P.~Yadav, M.~S. Steinbach, V.~Kumar, and G.~J. Simon, ``Mining electronic
  health records (ehrs): {A} survey,'' \emph{{ACM} Comput. Surv.}, vol.~50,
  no.~6, pp. 85:1--85:40, 2018.

\bibitem{moor2019early}
M.~Moor, M.~Horn, B.~Rieck, D.~Roqueiro, and K.~M. Borgwardt, ``Early
  recognition of sepsis with gaussian process temporal convolutional networks
  and dynamic time warping,'' in \emph{Proceedings of the Machine Learning for
  Healthcare Conference, {MLHC} 2019, 9-10 August 2019, Ann Arbor, Michigan,
  {USA}}, ser. Proceedings of Machine Learning Research, F.~Doshi{-}Velez,
  J.~Fackler, K.~Jung, D.~C. Kale, R.~Ranganath, B.~C. Wallace, and J.~Wiens,
  Eds., vol. 106.\hskip 1em plus 0.5em minus 0.4em\relax {PMLR}, 2019, pp.
  2--26.

\bibitem{aittokallio2010dealing}
T.~Aittokallio, ``Dealing with missing values in large-scale studies:
  microarray data imputation and beyond,'' \emph{Briefings Bioinform.},
  vol.~11, no.~2, pp. 253--264, 2010.

\bibitem{zhang2008parimputation}
S.~Zhang, ``Parimputation: From imputation and null-imputation to partially
  imputation,'' \emph{{IEEE} Intell. Informatics Bull.}, vol.~9, no.~1, pp.
  32--38, 2008.

\bibitem{raghunathan2001multivariate}
T.~E. Raghunathan, J.~M. Lepkowski, J.~Van~Hoewyk, P.~Solenberger
  \emph{et~al.}, ``A multivariate technique for multiply imputing missing
  values using a sequence of regression models,'' \emph{Survey methodology},
  vol.~27, no.~1, pp. 85--96, 2001.

\bibitem{chen2018neural}
T.~Q. Chen, Y.~Rubanova, J.~Bettencourt, and D.~Duvenaud, ``Neural ordinary
  differential equations,'' in \emph{Advances in Neural Information Processing
  Systems 31: Annual Conference on Neural Information Processing Systems 2018,
  NeurIPS 2018, December 3-8, 2018, Montr{\'{e}}al, Canada}, S.~Bengio, H.~M.
  Wallach, H.~Larochelle, K.~Grauman, N.~Cesa{-}Bianchi, and R.~Garnett, Eds.,
  2018, pp. 6572--6583.

\bibitem{wu2018sensor}
J.~Wu, Y.~Feng, and P.~Sun, ``Sensor fusion for recognition of activities of
  daily living,'' \emph{Sensors}, vol.~18, no.~11, p. 4029, 2018.

\bibitem{mitchell2007multi}
H.~B. Mitchell, \emph{Multi-sensor data fusion: an introduction}.\hskip 1em
  plus 0.5em minus 0.4em\relax Springer Science \& Business Media, 2007.

\bibitem{brena2020choosing}
R.~F. Brena, A.~A. Aguileta, L.~A. Trejo, E.~Molino{-}Minero{-}Re, and
  O.~Mayora, ``Choosing the best sensor fusion method: {A} machine-learning
  approach,'' \emph{Sensors}, vol.~20, no.~8, p. 2350, 2020.

\bibitem{diao2019data}
C.~Diao and B.~Wang, ``Data fusion of multivariate time series: Application to
  noisy 12-lead {ECG} signals,'' \emph{CoRR}, vol. abs/1803.01488, 2018.

\bibitem{zhu2019improvement}
W.~Zhu and F.~Xiao, ``Improvement of time series data fusion based on evidence
  theory and {DEMATEL},'' \emph{{IEEE} Access}, vol.~7, pp. 81\,397--81\,406,
  2019.

\bibitem{diao2017data}
C.~Diao, B.~Wang, and N.~Cai, ``Data fusion of multivariate time series based
  on local weighted zero-order prediction algorithm,'' in \emph{2017 36th
  Chinese Control Conference (CCC)}.\hskip 1em plus 0.5em minus 0.4em\relax
  IEEE, 2017, pp. 5663--5667.

\bibitem{harris1998multi}
C.~Harris, A.~Bailey, and T.~Dodd, ``Multi-sensor data fusion in defence and
  aerospace,'' \emph{The Aeronautical Journal}, vol. 102, no. 1015, pp.
  229--244, 1998.

\bibitem{maas2019facebook}
P.~Maas, S.~Iyer, A.~Gros, W.~Park, L.~McGorman, C.~Nayak, and P.~A. Dow,
  ``Facebook disaster maps: Aggregate insights for crisis response {\&}
  recovery,'' in \emph{Proceedings of the 16th International Conference on
  Information Systems for Crisis Response and Management, Val{\`{e}}ncia,
  Spain, May 19-22, 2019}, Z.~Franco, J.~J. Gonz{\'{a}}lez, and J.~H.
  Can{\'{o}}s, Eds.\hskip 1em plus 0.5em minus 0.4em\relax {ISCRAM}
  Association, 2019.

\bibitem{kreuter2020partnering}
F.~Kreuter, N.~Barkay, A.~Bilinski, A.~Bradford, S.~Chiu, R.~Eliat, J.~Fan,
  T.~Galili, D.~Haimovich, B.~Kim \emph{et~al.}, ``Partnering with a global
  platform to inform research and public policy making,'' in \emph{Survey
  Research Methods}, vol.~14, no.~2, 2020, pp. 159--163.

\end{thebibliography}

\end{document}